%% file: iclr2026_conference.tex
\documentclass{article}
\usepackage{iclr2026_conference}
\usepackage{times}
\input{math_commands.tex}

\usepackage{graphicx}
\usepackage{amsmath, amssymb, amsfonts}
\usepackage{dsfont}
\usepackage{subcaption}
\usepackage{booktabs}
\usepackage{tabularx}
\usepackage{xcolor}
\usepackage{mathtools}
\usepackage{cite}
\usepackage{hyperref}
\usepackage{url}
\usepackage{algorithm}
\usepackage{algorithmic}
 \usepackage{tikz}
 \usepackage{booktabs}
\usepackage{tabularx}
\usepackage{multirow}
\usepackage{makecell}
 \usetikzlibrary{positioning,arrows.meta,fit,calc}
\newtheorem{definition}{Definition}

\title{Are Large Language Models Good In-Context Learners for Financial Volatility Forecasting?}

\title{Regime-aware Financial Volatility Forecasting via In-Context Learning}

\author{
Saba Asaad$^{1}$, Shayan Mohajer Hamidi$^{2}$,  Ali Bereyhi$^{1}$ \\
$^{1}$Department of Electrical and Computer Engineering, University of Toronto, Toronto, Canada \\
$^{2}$Department of Electrical Engineering, Stanford University, Stanford, CA, USA \\
\texttt{saba.asaad@utoronto.ca, smohajer@stanford.edu, ali.bereyhi@utoronto.ca}
}

\begin{document}
\iclrfinalcopy
\maketitle

\begin{abstract}
This work introduces a regime-aware in-context learning framework that leverages large language models (LLMs) for financial volatility forecasting under nonstationary market conditions. The proposed approach deploys pretrained LLMs to reason over historical volatility patterns and adjust their predictions without parameter fine-tuning. We develop an oracle-guided refinement procedure that constructs regime-aware demonstrations from training data. An LLM is then deployed as an in-context learner that predicts the next-step volatility from the input sequence using demonstrations sampled conditional to the estimated market label. This conditional sampling strategy enables the LLM to adapt its predictions to regime-dependent volatility dynamics through contextual reasoning alone. Experiments with multiple financial datasets show that the proposed regime-aware in-context learning framework outperforms both classical volatility forecasting approaches and direct one-shot learning, especially during high-volatility periods.

% e show that naive in-context learning struggles to generalize across heterogeneous market regimes, particularly during periods of elevated volatility and rapid structural change.
% To address this limitation, 
% sampling strategies that select context examples based on inferred market regimes. 
%offers a promising new direction for volatility forecasting without retraining or gradient-based adaptation.
\end{abstract}
\section{Introduction}
Forecasting financial market volatility plays a central role in quantitative finance, with broad applications in risk management, derivative pricing, and portfolio allocation \citep{alexander2008market, andersen1998answering}. Volatility forecasting aims to estimate the dispersion of asset returns and serves as a key indicator of market uncertainty. Despite decades of research, accurate volatility forecasting remains a challenging problem \citep{Vol_Review}. This challenge is rooted in the dynamic nature of financial markets. Asset prices often fluctuate rapidly and abruptly, driven by a complex interplay of macroeconomic conditions, policy decisions, and geopolitical events. The interaction of these factors results in complex volatility dynamics, characterized by heavy-tailed return distributions and abrupt regime shifts \citep{rahimikia2021realised, abu1996introduction}. In this work, we investigate how large language models (LLMs) can adapt to such complex market dynamics when deployed as \textit{in-context learners} for volatility forecasting.
\paragraph*{Volatility forecasting models.}
Traditional approaches for volatility forecasting rely on econometric models or domain-specific machine learning techniques. However, their performance often deteriorates precisely during periods of elevated market stress \citep{Vol_Review}. Machine learning methods have been widely applied to volatility forecasting. These methods model the underlying task as a sequence prediction problem, in which large historical datasets are leveraged to learn long-term temporal patterns and complex dependencies in market data \citep{LSTM1,LSTM2,Trans}. Experiments show that these approaches are often effective under relatively \textit{stable} market conditions, where past observations provide a reasonable proxy for future behavior \citep{nie2022time}.
In practice, however, financial markets are highly nonstationary. As market conditions evolve, structural changes and unexpected shocks can quickly reduce the relevance of historical data, limiting the ability of the learning model to generalize to new environments \citep{geertsema2022measuring}. As a result, sequence models often struggle to adapt to evolving volatility regimes, particularly during turbulent periods characterized by abrupt increases in uncertainty \citep{MLReview}.

\paragraph*{In-context learning.}
Recently, LLMs have exhibited significant capabilities that extend beyond traditional natural language tasks. This includes the ability to reason over numerical information and perform sequence-based prediction from  historical context \citep{hwang2025decision, mahdavi2025integrating, li2025large}. Unlike conventional sequence models that rely on task-specific training or fine-tuning, LLMs can adapt their behavior during inference by conditioning on structured prompts and example demonstrations \citep{wang2023large, luo2024context}. This introduces a new paradigm, often referred to as \textit{in-context learning} \citep{dong2024survey}.

\subsection{Motivation and Contributions}
Considering the nonstationary and \textit{regime-dependence} nature of volatility forecasting, a natural question is raised: \textit{can LLMs act as effective in-context learners for financial volatility forecasting?} To answer this question, we develop an LLM-based volatility forecasting pipeline and investigate its performance in various market dynamics. We show that naive in-context learning faces similar limitations as those in conventional volatility forecasting approaches: while LLMs can produce reasonable forecasts during low-volatility periods, their predictions degrade substantially in high-volatility regimes, where large market movements and rapid structural changes dominate. Recognizing this \textit{regime-dependent} failure mode, we propose a \textit{regime-aware} in-context learning framework for realized volatility forecasting and investigate its characteristic in different market dynamics. 

Our key contributions are as follows: (i) we develop an oracle-guided refinement framework that deploys a fixed pretrained LLM as a foundation model to construct a \textit{regime-aware} demonstration for each training sample. The proposed iterative scheme leverages labeled data to extract informative context from the training set without any parameter fine-tuning. (ii) Using the generated context, we propose two \textit{regime-aware} conditional sampling strategies for volatility forecasting via in-context learning. These strategies select demonstrations conditional on an estimate of input regime label, enabling the LLM to adapt its predictions to changing market conditions. (iii) We evaluate the proposed framework on multiple financial datasets and compare it against standard classical volatility forecasting baselines. Our empirical results show that, while reflecting the well-known trade-off between \textit{regime adaptation} and \textit{average forecasting error}, regime-aware in-context learning can substantially outperform classical models. On the S\&P500 dataset, overall forecasting error is reduced by approximately 27\% relative to the best classical baseline.

\subsection{Related Work}
% Time series forecasting has progressed from classical models,
% Generalized auto-regressive conditional heteroskedasticity (GARCH) and exponential GARCH (EGARCH) are classical models for modeling time-varying conditional volatility 
Classical econometric approaches to volatility forecasting are commonly represented by generalized auto-regressive conditional heteroskedasticity (GARCH) type models \citep{franses1996forecasting, hansen2016exponential, nelson1991conditional}. Deep learning approaches to volatility forecasting train directly sequence models, such as recurrent networks and Transformers \citep{chen2015lstm, gamboa2017deep, wang2022ngcu}. As task-specific learners, these sequence models struggle to adapt to nonstationary environments and abrupt regime changes without retraining \citep{kong2025deep, ahmed2023transformers}. Pretrained LLMs, as a practical realization of foundation models, can perform numerical time series forecasting via in-context learning \citep{liang2024foundation}. 
\citet{gruver2023large} shows that pretrained LLMs can be directly prompted for time series prediction without task-specific fine-tuning by framing the task as a next-token prediction in a text space. This indicates that LLMs can match or even surpass classical time series models through effective tokenization and prompt design \citep{gruver2023large,jin2024timellm}. The experimental results highlight the effectiveness of pretrained self-attention mechanisms in LLMs for general sequential modeling \citep{zhou2023one}.

More recent efforts explore the integration of textual information with numerical time series to leverage the semantic reasoning capabilities of LLMs \citep{zheng2025fusing}. \citet{wang2024news} incorporate social media and news events into forecasting by aligning textual content with numerical fluctuations. The study deploys LLM-based agents to filter irrelevant information and perform high-level reasoning, leading to improved predictive accuracy. Other approaches propose multi-level alignment frameworks that couple pretrained language models with time series encoders to enhance predictive performance \citep{zhao2025enhancing}. 

% Recently, i
%emerged as a powerful framework that enables LLMs to adapt to new tasks by conditioning on a small number of task-specific examples at inference time, without any parameter updates or fine-tuning \citep{dong2024survey}. It has 
In-context learning has been successfully applied across diverse domains, demonstrating that pretrained models can infer task structure directly from contextual information \citep{dong2024survey}. The approach has shown notable learning capabilities, when combined with structured prompting techniques such as chain-of-thought reasoning \citep{zheng2025curse}. Despite this growing body of work, the application of in-context learning to time series forecasting remains relatively limited. Existing studies largely focus on settings with stable or homogeneous temporal dynamics \citep{jozi2022contextual}. The potential of in-context learning for financial volatility analysis, where abrupt regime shifts and nonstationary behavior are fundamental characteristics, has not been systematically explored. This gap motivates the development of in-context learning approaches that explicitly account for regime-dependent volatility dynamics.

% indicates that  is insufficient for volatility forecasting and that effective adaptation requires exposing the model to representative correction patterns across diverse market conditions.

% In this work, we propose a regime-aware in-context learning framework for realized volatility forecasting using fixed pre-trained LLMs. During training, our approach employs oracle-guided refinement to construct a high-quality demonstration pool, in which each example is explicitly labeled according to its volatility regime. At test time, the prevailing market regime is inferred from recent historical data, and regime-matched demonstrations are selected and provided as in-context examples to guide the LLM’s prediction. Importantly, this entire procedure operates without parameter fine-tuning, gradient-based updates, or oracle access during evaluation.

\color{black}

%\section{Related Work}
%\newpage

\section{Problem Formulation}
\label{sec:problem}
We study the problem of volatility forecasting for financial assets using historical price observations. Given a univariate financial time series, the objective is to predict future market volatility, as measured by realized variance, based on past observations. To formulate this objective, let us first define the \textit{log return} as a measure of retuen.
\begin{definition}[Log return]
    Let $P_t$ denote the daily closing price of a financial asset at time $t$.
The corresponding daily log return is defined as
\begin{equation}
r_t = \ln\left(\frac{P_t}{P_{t-1}}\right).
\end{equation}
\end{definition} 

In general, volatility describes the variability of returns. Following standard practice in volatility forecasting, we use the \textit{squared daily log return} as a proxy for daily realized variance, i.e.
\begin{equation}
\nu_t = r_t^2,
\end{equation}
and consider it as a measure for volatility. 

The volatility forecasting can be modeled as a sequence prediction task: at each time step $t$, we construct a fixed-length historical input sequence using a sliding window of size $w$, i.e. 
\begin{equation}
\mathbf{x}_t =
\left[
(r_{t-w+1}, \nu_{t-w+1}), \ldots, (r_t, \nu_t)
\right].
\end{equation}
The sequence $\mathbf{x}_t$ captures both recent returns and volatility patterns. The task is to predict the next realized variance $\nu_{t+1}$ given the historical input $\mathbf{x}_t$. In this work, we aim to induce this predictor via prompting a pretrained LLM, without any parameter fine-tuning or gradient-based updates.

Using the above definitions, the design task is formulated as follows: Let
\begin{equation}
    \mathcal{C}=\left\lbrace (\mathbf{x}_t, \nu_{t+1}): t \in \lbrace 0,\ldots, n-1 \rbrace\right\rbrace
\end{equation}
be a dataset built from historical log return samples. Assume that a fixed pre-trained LLM $\mathcal{M}$ is given. The goal is to induce a mapping $f: \mathbf{x}_t \mapsto \hat{\nu}_{t+1}$, from $\mathcal{C}$ by prompting $\mathcal{M}$. %such that the prediction $\widehat{\nu}_{t+1}$ generalizes with respect to some %closely matches the ground-truth realized variance $\nu_{t+1}$.

\section{Methodology} 
% We deploy a pre-trained LLM, without any parameter fine-tuning or gradient-based updates, to build a volatility predictor. 
The proposed design uses an oracle-based refinement loop to construct a demonstration pool. The pool is then used to forecast the realized variance via in-context learning. %In the sequel, we present the design in detail. 

\begin{figure}
    \centering
    \includegraphics[width=1\linewidth]{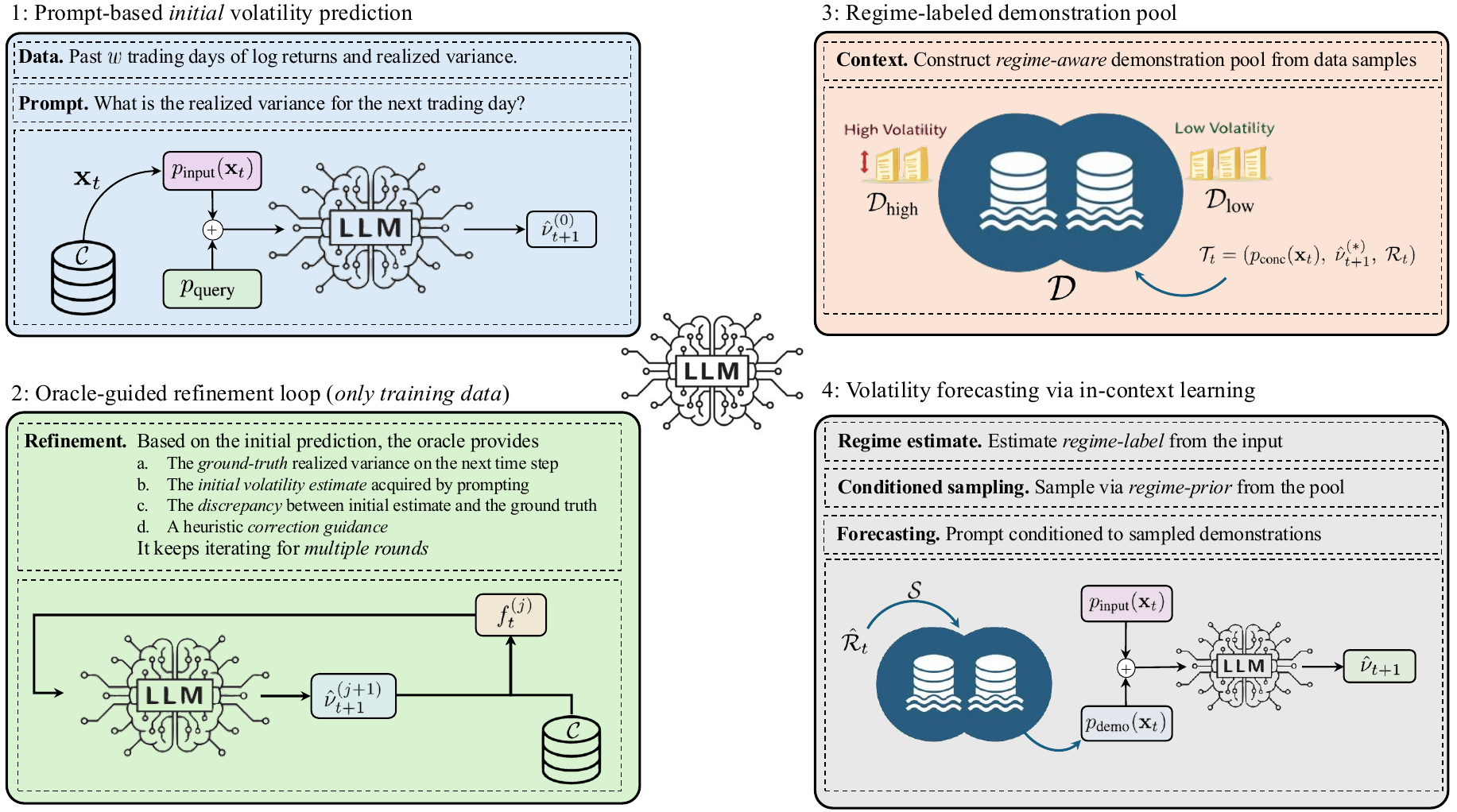}
    \caption{Illustration of the proposed prompt-based volatility forecasting framework.
(1) A fixed pretrained LLM is first prompted with recent log returns and realized variance to produce an initial estimate. (2) Oracle feedback is build from the ground truth and used iteratively to refine predictions. (3) A regime-labeled demonstration pool is constructed from refined predictions. (4) Demonstrations are selected conditional to estimated volatility regime and provided as in-context examples to generate a forecast of the next-day realized variance.}
    \label{fig:diagram}
\end{figure}

\subsection{Initial Prompt-based Volatility Prediction}
% We first formulate volatility forecasting as a prompt-based numerical prediction task using the fixed pre-trained LLM $\mathcal{M}$. A key step in this setting is to 
% We represent numerical market dynamics in a form that is effectively processed by LLMs. To address this, 
We transform recent numerical time-series observations into structured natural language prompts. This enables the LLM to condition on and reason over historical volatility patterns through in-context learning. Specifically, given a historical input sequence $\mathbf{x}_t$ at time $t$ consisting of the most recent $w$ days, we represent the sequence as a structured textual prompt, denoted by $p_{\text{input}}(\mathbf{x}_t)$, which encodes log returns and realized variance values in chronological order. In addition, we provide a task-specific query prompt $p_{\text{query}}$ that specifies the forecasting objective, including the prediction window and the target variable. %In our setting, the query prompt instructs the LLM to predict the next-day realized variance based on the provided historical context. 
Given the prompts, the LLM generates the initial volatility predictor
\begin{equation}
\hat{\nu}_{t+1}^{(0)}
=
\mathcal{M} \left( p_{\text{input}} (\mathbf{x}_t) \oplus p_{\text{query}} \right), \label{eq:Prompt}
\end{equation}
where %$\mathcal{M}(\cdot)$ denotes a fixed pre-trained LLM and 
$\oplus$ denotes prompt concatenation. %We denote the current response by $\widehat{\nu}_{t+1}$ with $\widehat{\nu}_{t+1}= \widehat{\nu}_{t+1}^0$ in the first refinement round, and update it to the refined response in subsequent rounds.

The LLM $\mathcal{M}$ operates in an inference-only mode. Thus, all predictive behavior arises solely from prompt design and in-context learning. While this direct prompting mechanism in (\ref{eq:Prompt}) is sufficient to produce numerical forecasts, it does not consistently capture the scale and variability inherent in financial volatility time series. In fact, the initial LLM forecast is often imprecise, particularly for financial volatility time series with heavy tails and regime shifts. In the sequel, we use this initial prediction to construct a \textit{refinement-based} and regime-aware demonstration pool, which enables us to predict the volatility via in-context learning. % introduced in the following sections.

\subsection{Constructing Demonstration Pool via Oracle-Guided Refinement}
\label{sec:refinement}
We leverage oracle feedback for demonstration construction to expose the model to explicit correction signals and thereby construct higher-quality in-context examples. We then collect these examples in a regime-aware pool for in-context learning.

\paragraph{Demonstration construction mechanism.}
Starting with the prediction in (\ref{eq:Prompt}) as $\hat{\nu}_{t+1}^{(0)}$, for a given response $\hat{\nu}_{t+1}^{(j)}$ and the feedback $f_t^{(j)}$, the LLM performs a refinement step via
\begin{equation}
\hat{\nu}_{t+1}^{(j+1)}
=
\mathcal{M}\left(\hat{\nu}_{t+1}^{(j)},\, f_t^{(j)}\right),
\end{equation}
where the feedback incorporates the ground-truth realized variance, the current prediction, the resulting prediction error, and a heuristic correction signal. Specifically, we define
\begin{equation}
f_t^{(j)}
=
\left(
\nu_{t+1},
\hat{\nu}_{t+1}^{(j)},
e_t^{(j)},
\mathcal{H}_t^{(j)}
\right),
\end{equation}
where $e_t^{(j)} = \vert \hat{\nu}_{t+1}^{(j)} - \nu_{t+1} \vert$ is the prediction error, and $\mathcal{H}_t^{(j)}$ denotes a simple heuristic suggestion, e.g., encouraging adjustment toward a recent volatility average. This feedback-driven refinement loop allows the LLM to explicitly reason about its prediction errors and perform structured test-time numerical correction. Importantly, oracle access to the ground truth is used only during demonstration construction and is never available at test time.

After a fixed number of refinement iterations, we retain the final refined prediction $\hat{\nu}_{t+1}^{(*)}$ and store each demonstration as a tuple
\begin{equation}
\mathcal{T}_t = (
p_{\text{conc}} (\mathbf{x}_t) ,\;
\hat{\nu}_{t+1}^{(*)},\;
\mathcal{R}_t),    
\end{equation}
where $p_{\text{conc}} (\mathbf{x}_t) = p_{\text{input}} \oplus p_{\text{query}}$ denotes the history prompt text, and $\mathcal{R}_t$ is the \textit{volatility regime label} determined by the next-day realized variance according to
\begin{equation}
\mathcal{R}_t =
\begin{cases}
\text{high} & \nu_{t+1} \ge \tau \\
\text{low} & \text{otherwise}
\end{cases},
\end{equation}
for some threshold $\tau$ computed exclusively from the training data. 

\paragraph{Regime-aware demonstration pool.}
We collect the demonstration samples $\mathcal{T}_t$ in a demonstration pool, i.e. %we build %$\mathcal{D}$ as
% \begin{equation}
$\mathcal{D}=\left\lbrace \mathcal{T}_t: t \in \lbrace 0,\ldots, n-1 \rbrace\right\rbrace$.
% \end{equation}
The volatility regime label $\mathcal{R}_t$ yields two conditional pools: a high-volatility pool $\mathcal{D}_{\text{high}}$, whose demonstration samples are labeled \textit{high}, i.e.
\begin{equation}
    \mathcal{D}_{\text{high}}=\left\lbrace \mathcal{T}_t\in \mathcal{D}:  \mathcal{R}_t = \text{high}\right\rbrace,
\end{equation}
and a low-volatility pool $\mathcal{D}_{\text{low}}$, whose demonstration samples are labeled \textit{low}, i.e.
\begin{equation}
    \mathcal{D}_{\text{low}}=\left\lbrace \mathcal{T}_t\in \mathcal{D}:  \mathcal{R}_t = \text{low}\right\rbrace.
\end{equation}
Note that the pool is constructed from training data and remains unchanged throughout evaluation.

\subsection{Forecasting via In-Context Learning}
\label{sec:regime_inference}
Using the regime-aware demonstration pool, we now deploy in-context learning to estimate the next realized variance from selected demonstration samples. In the sequel, we illustrate demonstration selection strategies. Motivated by our regime-labeled demonstration pool, we  propose two \textit{regime-aware selection} strategies. % affect the ability of LLMs to perform volatility forecasting via in-context learning. To this end, we evaluate several demonstration selection strategies, including our proposed regime-aware approach.

\subsubsection{Demonstration Sampling}
\label{DemoSamp}
\paragraph{Random selection baseline.}
A widely-used baseline selection approach is \textit{random} selection, where we uniformly sample $K$ demonstration tuples from the full demonstration pool $\mathcal{D}$. Random selection is the simplest and most widely adopted strategy for in-context learning \citep{qin2024context,peng2024revisiting}. We deploy this strategy as a natural reference point for evaluating more structured selection methods. In addition, random selection can act as an implicit explorer, which  gives upper-bounds on performance gains achievable through demonstration-based prompting.
%%%%%%%%%%%%%%%%%%

\paragraph{Proposal I: conditional selection via fixed label prior.}
Given the labeled demonstration pool, a natural extension of random selection is to sample conditionally from labeled pools according to the label prior. This approach ensures mixing demonstrations from both regimes. Specifically, for a fixed label prior $\alpha \in [0,1]$, we sample $\lfloor \alpha K\rfloor$ demonstration tuples from $\mathcal{D}_{\text{low}}$ and $K- \lfloor \alpha K\rfloor$ tuples from $\mathcal{D}_{\text{high}}$. This label-conditioned sampling controls label diversity of the context without explicit conditioning on the test-time regime. Intuitively, this approach allows us to assess whether performance gains arise from regime awareness or simply from balanced exposure to different volatility regimes. With limited label skew in the demonstration pool, one can set $\alpha$ to the one estimated from the pool. In scenarios with highly skewed pools, the approach can benefit from tuning $\alpha$. %(In experiments we report results for a small set of $\alpha$ values; see Section~\ref{sec:experiments}.)
%%%%%%%%%%%%%%%%%%%%%%%%
\paragraph{Proposal II: conditional selection via  estimated label.}
In this strategy, the demonstration selection is conditioned on an estimated volatility regime at test time to directly tune the label prior from the date. In this approach, we first infer the current volatility regime from the test history using only information available up to time $t$: we compute a recent-volatility signal as the mean of the $m$ recent realized variances, i.e., from $\mathbf{x}_t$, we compute
\begin{equation}
   s_t= \frac{1}{m} \sum_{u=t-m+1}^{t} \nu_u ,
\end{equation}
for some $m<w$. We then estimate the current volatility regime as 
\begin{align}
    \hat{\mathcal{R}}_t = \begin{cases}
        \text{hight} &s_t \ge \tau'\\
        \text{low} &\text{otherwise}
    \end{cases},
\end{align}
% high volatility if $s_t \ge \tau$, and low volatility otherwise, 
for some $\tau'$, which can be generally different from $\tau$. Note that $s_t$ depends on information up to time $t$ and $\tau'$ is fixed prior to evaluation. The regime detection procedure is hence leakage-free. Based on the estimated regime $\hat{\mathcal{R}}_t$, we sample the demonstration pool conditionally with a prior: for $\alpha_{\rm low} > \alpha_{\rm high}$, we %first compute the estimate $\hat{\mathcal{R}}_t$ and then 
sample $K$ tuples $\mathcal{D}$ conditionally (Proposal I) with $\alpha_{\hat{\mathcal{R}}_t}$. %we randomly sample $K$ demonstrations from $\mathcal{D}_{\text{high}}$; otherwise, we sample $K$ demonstrations from $\mathcal{D}_{\text{low}}$. Sampling is performed without replacement.
%%%%%%%%%%%%
\subsubsection{Forecasting}
Let $\mathcal{S}$ denote the sampling approach. The demonstration is sampled as $p_{\text{demo}}(\mathbf{x}_t)  = \mathcal{S}(\mathbf{x}_t)$. The LLM is then prompted to generate a single-shot forecast of the next realized variance, i.e.
\begin{equation}
\hat{\nu}_{t+1}
=
\mathcal{M}\left(
p_{\text{demo}}(\mathbf{x}_t)  \;\oplus\; p_{\text{input}}(\mathbf{x}_t)
\right).
\end{equation}
% where $p_{\text{demo}}(t)$ denotes the concatenation of the selected regime-matched demonstrations.

Note that no oracle feedback, ground-truth information, or refinement loop is used during evaluation. All adaptation to changing volatility regimes arises solely from regime-aware demonstration selection and in-context learning within a single prompting of the LLM.
%Crucially, these demonstrations do not merely provide numerical input--output pairs.Instead, they expose the LLM to calibrated response patterns that implicitly encode how volatility forecasts should react under different market regimes.Demonstrations drawn from low-volatility periods encourage smooth and stable predictions, whereas demonstrations from high-volatility periods illustrate when larger corrective adjustments are appropriate.By constructing the demonstration set to cover both regimes, the LLM is guided to balance stability during calm periods with responsiveness during volatility spikes.

%Experiment
\section{Experiments} 
We evaluate our proposed refined-LLM framework through multiple experiments and compare it against the benchmark approaches for volatility forecasting. %The experiments are designed to assess both effectiveness and scalability of the proposed scheme.

\paragraph{Data.}
% \label{subsec:datasets}
To evaluate the robustness of the proposed framework, we consider multiple financial datasets. Namely, we consider the following datasets:
%including equity indices and foreign exchange rates. These datasets are given below. %Studying this wide range of assets enables a comprehensive assessment of volatility forecasting performance across diverse market conditions. The detail of the considered datasets is given below. 
\begin{itemize}
\item[(i)] \textbf{S\&P500}
We use daily price data of the S\&P500 index obtained from the Stooq financial database. 
\item[(ii)]
\textbf{NASDAQ.}
To evaluate performance in a different equity market segment, we include daily price data of the NASDAQ Composite index, also sourced from Stooq.
\item[(iii)]
\textbf{Foreign exchange.}
We consider the EUR-to-USD exchange rate using daily closing prices. The foreign exchange market provides a contrasting volatility structure due to its high liquidity and continuous trading \citep{FXBook}.
\end{itemize}

% \paragraph{Cryptocurrency.}
% We include Bitcoin (BTC) daily realized variance to evaluate performance in a highly volatile and non-stationary market. Daily realized variance is constructed from BTC price data following the same procedure used for the other assets.

Across all datasets, we adopt a consistent one-day-ahead forecasting setup and apply identical preprocessing and evaluation protocols to ensure fair comparisons. Daily realized variances are computed from log returns and serve as the prediction target in the one-day-ahead forecasting task. The dataset is split  into 70\% training data, i.e. used for demonstration pool construction and in-context learning, and 30\% test data. Refinement and demonstration construction are performed exclusively on the training set.

\paragraph{Baseline.}
% \label{subsec:baselines}
We consider a one-day-ahead volatility forecasting task, where the objective is to predict the next-day realized variance using a fixed-length window of historical observations.
Our approach employs a pre-trained LLM as a foundation model and operates in a purely prompt-based manner, i.e., without any parameter fine-tuning or gradient-based updates. Unless otherwise stated, we use \texttt{gpt-4o-mini} as the underlying foundation model. To evaluate the effectiveness of the proposed refinement-based in-context learning scheme, we compare our proposals against several representative baseline methods commonly used in volatility forecasting \citep{RVBook,GARCH-Models}.

\begin{enumerate}
\item \textbf{Rolling mean.}
Rolling mean is a well-known volatility forecasting baseline, which predicts the next-day realized variance as the average of the most recent realized variance observations \citep{RV3,RVBook}.
\item \textbf{HAR.}
The heterogeneous autoregressive (HAR) model is an alternative baseline for realized variance prediction. HAR models log-realized variance using daily, weekly, and monthly components, capturing volatility persistence across multiple time scales \citep{HAR-RV}.
\item \textbf{GARCH(1,1).}
GARCH refers to a class of models, which estimate conditional variance based on past squared returns and past variance \citep{GARCH-Models}. We consider GARCH(1,1) as the baseline, which serves as a standard benchmark in financial econometrics \citep{GARCH1,GARCH2}.
\item \textbf{GJR-GARCH.}
This model extends GARCH by incorporating \textit{asymmetric effects} to capture the leverage effect, whereby negative returns lead to higher subsequent volatility \citep{GJR-GARCH}. GJR-GARCH approach enables the model to better capture nonlinear and asymmetric responses to shocks in the data.
\item \textbf{One-shot learning.} In this baseline, the LLM directly predicts next-day realized variance from the input window using a single prompt, without refinement or demonstrations.
\end{enumerate}

\paragraph{Evaluation metrics.}
% \label{subsec:metrics}
We use standard regression-based error metrics. Specifically, we report the root mean squared error (RMSE)
% , i.e., 
% \begin{equation}
%     \mathrm{RMSE} = \sqrt{\frac{1}{v} \sum_{t=1}^{v} ( \hat{\nu}_t - \nu_t )^2},
% \end{equation}
and the mean absolute error (MAE).
% , i.e.
% \begin{equation}
%     \mathrm{MAE} = \frac{1}{v} \sum_{t=1}^{v} | \hat{\nu}_t - \nu_t |,
% \end{equation}
% with $v$ denoting the test size. 
% which are widely used in volatility forecasting.
% Table~\ref{tab:metrics} summarizes the definitions of these metrics, where 
% In these definitions, $\nu_t$ denotes the realized variance at time $t$, and $\hat{\nu}_t$ is the prediction. %, and $N$ is the total number of forecasts.
% \begin{table}[h]
% \centering
% \caption{Definitions of evaluation metrics}
% \label{tab:metrics2}
% \renewcommand{\arraystretch}{1.1}
% \begin{tabular}{l c}
% \hline
% \textbf{Metric} & \textbf{Definition} \\
% \hline
% MAE &
% $\frac{1}{N} \sum\nolimits_{t=1}^{N} | \hat{\nu}_t - \nu_t |$ \\
% RMSE &
% $\sqrt{\frac{1}{N} \sum\nolimits_{t=1}^{N} ( \hat{\nu}_t - \nu_t )^2}$ \\
% \hline
% \end{tabular}
% \end{table}
Note that RMSE penalizes larger deviations more heavily, and thus emphasizes the impact of \textit{large forecasting errors}. MAE, on the other hand, treats all deviations linearly, and hence provides a \textit{balanced measure} of forecasting accuracy that is less sensitive to large errors.

In addition to overall MAE and RMSE, we report \textit{regime-wise MAE} to assess the forecasting performance under different volatility conditions: let $\tau$ be the $q$-quantile of realized variance in the training set. We partition the test set $\mathcal{C}_{\text{test}}$ into two subsets, namely
\begin{equation}
\mathcal{C}_{\text{low}} = \{ t \in \mathcal{C}_{\text{test}} : \nu_{t+1} < \tau \}, \qquad
\mathcal{C}_{\text{high}} = \{ t \in \mathcal{C}_{\text{test}} : \nu_{t+1} \ge \tau \}.
\end{equation}
%Let $\tau$ denote the volatility threshold computed from the training data, defined as the $q$-quantile of the realized variance in the training set:
%\begin{equation}
%\tau \;=\; \mathrm{Quantile}\!\left(\{\nu_t : t \in \mathcal{T}_{\text{train}}\},\, q \right),
%\end{equation}
%where $q \in (0,1)$ is a fixed quantile level. We partition the test set into two subsets:
The \textit{low-volatility MAE} ($\mathrm{MAE}_{\text{low}}$) and \textit{high-volatility MAE} ($\mathrm{MAE}_{\text{high}}$) are then defined as 
\begin{equation}
    \mathrm{MAE}_{i} = 
    \frac{1}{|\mathcal{C}_{i}|}
\sum_{t \in \mathcal{C}_{i}}
\left| \hat{\nu}_{t+1} - \nu_{t+1} \right|,
\end{equation}
for $i\in \lbrace\text{low}, \text{high}\rbrace$. 
% \begin{table}[h]
% \centering
% \caption{Regime-wise evaluation metrics}
% \label{tab:metrics}
% \renewcommand{\arraystretch}{1.1}
% \begin{tabular}{l c}
% \hline
% \textbf{Metric} & \textbf{Definition} \\
% \hline
% $\mathrm{MAE}_{\text{low}}$ &
% $\frac{1}{|\mathcal{T}_{\text{low}}|}
% \sum_{t \in \mathcal{T}_{\text{low}}}
% \left| \widehat{\nu}_{t+1} - \nu_{t+1} \right|$ \\
% $\mathrm{MAE}_{\text{high}}$ &
% $\frac{1}{|\mathcal{T}_{\text{high}}|}
% \sum_{t \in \mathcal{T}_{\text{high}}}
% \left| \widehat{\nu}_{t+1} - \nu_{t+1} \right|$ \\
% \hline
% \end{tabular}
% \end{table}
This regime-wise evaluation allows us to assess forecasting accuracy separately during calm periods and volatility spikes.

% \subsection{Experimental Results}

\paragraph{Implementation details and hyperparameters.}
%Unless otherwise stated, \texttt{gpt-4o-mini} is deployed with temperature being set to zero. 
Across all experiments, a fixed-length historical window of $w=7$ days is considered. Each input window consists of past log returns and realized variance values. During demonstration construction, we build a pool of $n=500$ refined demonstrations. The initial samples are randomly selected from training data and the demonstration tuples are constructed by the refinement loop after 3 iterations. 

During the evaluation, we sample $K=5$ in-context demonstration tuples for each prediction. When constructing the demonstration pool via conditional sampling with fixed prior, i.e. Proposals I and II in Section~\ref{DemoSamp}, we enforce the high-regime fraction, i.e. $\alpha$ factor, to ensure adequate representation of high-volatility periods. In the conditional selection approach with estimated label, i.e., Proposal II in Section~\ref{DemoSamp}, the volatility regime label is estimated from the recent history by averaging the last $m=3$ realized variance values. 
% using the last $m=3$ time samples, by averaging the most recent realized variance values. 
Volatility regimes are defined using a training-derived threshold $\tau$, calculated as the $q=0.8$ quantiles of the variance realized in the training set. 

%%%%
For all classical baselines, we follow standard configurations commonly adopted in the literature \citep{HAR-RV,ANDERSEN200143}.
The HAR model employs daily, weekly, and monthly components with window lengths of 1, 5, and 22 trading days, respectively.
For GARCH-based models, we use the canonical GARCH(1,1) and GJR-GARCH(1,1,1) specifications with Gaussian innovations.
All baseline hyperparameters are fixed \emph{a priori} and are not tuned on the test set.

%%%
% All experiments use a fixed random seed for reproducibility. 
% The iterative refinement procedure during demonstration construction is run for a maximum of $R=3$ refinement steps per demonstration. \color{black}

\paragraph{Experimental results.} Table~\ref{tab:multi_asset_results} reports one-day-ahead volatility forecasting performance across multiple assets, all normalized to $10^{-4}$. Classical baselines exhibit the expected \textit{regime-dependent} behavior. The rolling mean baseline performs worst overall, reflecting its inability to adapt to changes in volatility dynamics. HAR achieves highest accuracy during \textit{low-volatility} periods, following its long-memory structure \citep{HAR-RV,CLEMENTS2021106285,Ma15082019}. However, it incurs substantially larger errors during high-volatility episodes \citep{Ma15082019}. In contrast, GJR-GARCH consistently attains the lowest error in the high-volatility regime among classical models, confirming its effectiveness in capturing asymmetric volatility dynamics \citep{ENGLE}. This describes a known trade-off in volatility forecasting, i.e. achieving a better regime adaptation at the cost of higher average error \citep{Hansen,reading20448}.

In-context learning approaches exhibit a similar trade-off between regime adaptation and overall performance. The one-shot learning baseline, i.e., taking volatility prediction after the initial prompt, produces smooth forecasts that under-react to abrupt volatility increases. As a result, it shows the highest error among all LLM-based variants, particularly during high-volatility periods ($\mathrm{MAE}_{\text{high}}$). Introducing in-context demonstrations substantially improves performance: \textit{regime-agnostic} random demonstration selection already yields accuracy comparable to HAR. This indicates that the LLM is able to internalize volatility persistence and infer from historical examples without explicit parameter updates.

Incorporating regime information further enhances performance. The conditional demonstration sampling strategy with fixed prior (Proposal I) improves high-volatility accuracy ($\mathrm{MAE}_{\text{high}}$) while maintaining competitive performance during normal market conditions ($\mathrm{MAE}_{\text{low}}$). The conditional sampling approach with regime label estimate (Proposal II) demonstrates the same trend. It outperforms the other in-context learners, due to its adaptiveness. Among both classical and in-context learning approaches, Proposal II achieves the lowest overall MAE and RMSE.

Compared to classical approaches, regime-aware in-context learners (Proposals I and II) significantly reduce high-volatility error ($\mathrm{MAE}_{\text{high}}$). This \textit{regime adaptation} comes at the cost of higher error in normal market ($\mathrm{MAE}_{\text{low}}$), describing the similar trade-off as in GARCH-based models, but at an improved operating point: for S\&P500, the regime-aware in-context learner (Proposal II) reduces the high-volatility MAE by approximately $27\%$ compared to GJR-GARCH, i.e. the strongest classical baseline in this regime.

\begin{table*}[t]
\centering
\small
\setlength{\tabcolsep}{4pt} % tighter columns
\renewcommand{\arraystretch}{1.1} % a bit tighter rows
\caption{Volatility forecasting performance across multiple assets in terms of RMSE and MAE. The results are normalized to $10^{-4}$. Low- and high-volatility regimes are defined using the 80th percentile of realized variance. Regime-aware in-context learning enhances the performance, due to better estimate in \textit{high} regime.}
\label{tab:multi_asset_results}
\begin{tabularx}{\textwidth}{l l c c | c c r}
\toprule
Dataset & Method & $\mathrm{MAE} (\downarrow)$ & $\mathrm{RMSE} (\downarrow)$ & $\mathrm{MAE}_{\text{low}} (\downarrow)$ & $\mathrm{MAE}_{\text{high}} (\downarrow)$ &$\times10^4$ \\
\hline

\multirow{8}{*}{S\&P500}
& Rolling mean  & $1.41$ & $4.87$ & $0.73$ & $4.13$& \\
& HAR & $\mathbf{1.17}$ & $5.36$ & $\mathbf{0.25}$ & $4.85$& \\
& GARCH(1,1) & $1.34$ & $4.80$ & $0.71$ & $3.86$& \\
& GJR-GARCH & $1.33$ & $\mathbf{4.64}$ & $0.72$ & $\mathbf{3.76}$& \\
\cline{2-6}
& One-shot learning & $1.72$ & $6.68$ & $0.87$ & $5.09$& \\
& Random selection & $1.22$ & $4.77$ & $0.65$ & $3.12$& \\
& \textbf{Regime-aware (fixed prior)} & $1.18$ & $4.64$ & $\mathbf{0.61}$ & $3.09$& \\
& \textbf{Regime-aware (label estimate)} & $\mathbf{1.14}$ & $\mathbf{4.51}$ & $0.65$ & $\mathbf{2.73}$& \\

\midrule

\multirow{8}{*}{NASDAQ}
& Rolling mean & $2.12$ & $5.88$ & $1.21$ & $5.74$& \\
& HAR & $\mathbf{1.70}$ & $6.14$ & $\mathbf{0.44}$ & $6.75$& \\
& GARCH(1,1) & $2.00$ & $5.59$ & $1.22$ & $5.10$& \\
& GJR-GARCH & $1.97$ & $\mathbf{5.51}$ & $1.21$ & $\mathbf{5.01}$& \\
\cline{2-6}
& One-shot learning & $2.32$  & $7.06$ & $1.37$  & $6.13$&  \\
& Random selection & $1.80$ & $5.13$ & $0.92$ & $5.31$& \\
& \textbf{Regime-aware (fixed prior)} & $1.72$ & $5.09$ & $\mathbf{0.89}$ & $5.08$ & \\
& \textbf{Regime-aware (label estimate)} & $\mathbf{1.67}$ & $\mathbf{5.05}$ & $0.90$ & $\mathbf{4.75}$& \\

\midrule

\multirow{8}{*}{EUR/USD}
& Rolling mean  & $0.27$ & $0.54$ & $0.18$ & $0.64$& \\
& HAR & $\mathbf{0.20}$ & $0.50$ & $\mathbf{0.07}$ & $0.76$& \\
& GARCH(1,1) & $0.26$ & $0.47$ & $0.20$ & $\mathbf{0.52}$& \\
& GJR-GARCH & $0.26$ & $\mathbf{0.47}$ & $0.20$ & $0.53$& \\
\cline{2-6}
& One-shot learning & $0.37$  & $0.63$  & $0.28$ & $0.76$&  \\
& Random selection & $0.24$ & $0.49$ & $0.18$ & $0.48$& \\
& \textbf{Regime-aware (fixed prior)} & $0.21$ & $0.47$ & $\mathbf{0.16}$ & $0.45$& \\
& \textbf{Regime-aware (label estimate)} & $\mathbf{0.19}$ & $\mathbf{0.45}$ & $0.17$ & $\mathbf{0.31}$& \\

\bottomrule
\end{tabularx}
\end{table*}

\section{Conclusions}
We developed a regime-aware in-context learning framework for volatility forecasting using fixed pre-trained large language models and evaluated its performance on real-world financial data. Our approach employs an oracle-guided refinement procedure to construct a demonstration pool in which each example is explicitly labeled by volatility regime. At inference, the LLM is guided by regime-matched demonstrations selected through conditional sampling from this pool. By incorporating explicit regime information into the in-context learning process, the proposed framework effectively balances the stability of long-memory models with the adaptive response required under asymmetric market conditions. Empirical results demonstrate that this regime-aware in-context learning strategy yields consistent improvements in both overall and regime-specific accuracy, particularly during high-volatility periods. Exploring retrieval-based demonstration selection for enhanced context extraction is an interesting direction for future work. 

% We developed a regime-aware in-context learning framework for volatility forecasting using pretrained LLMs and studied its performance on real-world financial data. Our approach employs an oracle-guided refinement procedure to construct a demonstration pool whose examples are explicitly labeled by volatility regime. During inference, the LLM is guided by regime-matched demonstrations selected by conditional sampling from the demonstration pool. The incorporation of explicit regime information  enables the proposed framework to effectively balance between stable long-memory and the adaptive response to asymmetric market conditions. Empirical results show that the proposed regime-aware in-context learning strategy yields consistent improvements in both overall accuracy and regime-specific performance, particularly during high-volatility periods.
% The work in this direction is currently ongoing. 

\bibliography{iclr2026_conference}
\bibliographystyle{iclr2026_conference}

\appendix

% in preamble:
% \usepackage{tikz}
% \usetikzlibrary{positioning,arrows.meta,fit,calc}

\end{document}

%% file: math_commands.tex
\usepackage{amsmath,amsfonts,bm}

\def\eqref#1{equation~\ref{#1}}
\def\1{\bm{1}}

\DeclareMathAlphabet{\mathsfit}{\encodingdefault}{\sfdefault}{m}{sl}
\SetMathAlphabet{\mathsfit}{bold}{\encodingdefault}{\sfdefault}{bx}{n}